\pdfoutput=1

\documentclass[letterpaper, 10 pt, conference]{ieeeconf}
\IEEEoverridecommandlockouts
\newif\ifanonymous
\anonymousfalse

\usepackage[T1]{fontenc}
\usepackage[utf8]{inputenc}
\usepackage{graphicx}
\usepackage{amsmath,amssymb}
\usepackage{booktabs}
\usepackage{multirow}
\usepackage{siunitx}
\usepackage{microtype}
\usepackage{xcolor}
\usepackage{tikz}
\usepackage{balance}
\usepackage[hidelinks]{hyperref}
\usepackage[capitalize,noabbrev]{cleveref}

\usetikzlibrary{arrows.meta,positioning,calc,fit,backgrounds}

\graphicspath{{figures/}}

\definecolor{nfgink}{HTML}{0B0B0B}
\definecolor{nfgink2}{HTML}{52514E}
\definecolor{nfgblue}{HTML}{2A78D6}
\definecolor{nfggreen}{HTML}{1BAF7A}
\definecolor{nfgorange}{HTML}{D1653A}
\definecolor{nfgpurple}{HTML}{8A6FD1}
\definecolor{nfgred}{HTML}{C2553F}

\newcommand{\DemosPick}{150}

\newcommand{\DemosStack}{1000}

\newcommand{\FlagDinoValid}{+24.7}
\newcommand{\FlagDinoValidP}{0.004}

\newcommand{\FlagDinoValue}{+6.2}

\newcommand{\FlagMogeValid}{+34.1}

\newcommand{\FlagMogeValue}{+0.7}

\newcommand{\FlagRnValid}{+22.9}

\newcommand{\FlagRnValue}{+12.2}
\newcommand{\FlagRnValueP}{5{\times}10^{-4}}

\newcommand{\GainAbsMax}{4.1}

\newcommand{\GainN}{26}

\newcommand{\HalfRelDino}{6.2}

\newcommand{\HalfRelMoge}{12.6}

\newcommand{\HalfRelRn}{0.9}

\newcommand{\HalfRelStackRn}{1.2}

\newcommand{\HalfRelSwapAnyDino}{1.9}
\newcommand{\HalfRelSwapAnyDinoHi}{2.9}
\newcommand{\HalfRelSwapAnyDinoLo}{1.5}

\newcommand{\HalfRelSwapAnyMoge}{6.1}
\newcommand{\HalfRelSwapAnyMogeHi}{9.5}
\newcommand{\HalfRelSwapAnyMogeLo}{1.5}

\newcommand{\HalfRelSwapAnyRn}{0.9}
\newcommand{\HalfRelSwapAnyRnHi}{7.6}
\newcommand{\HalfRelSwapAnyRnLo}{0.4}

\newcommand{\HalfRelSwapAnyStackRn}{1.1}

\newcommand{\HoleGlobal}{2.1}
\newcommand{\HoleWrist}{16.4}

\newcommand{\KWrist}{1.79}

\newcommand{\MissWristEighthH}{2.0}

\newcommand{\MissWristFourH}{64.3}

\newcommand{\MissWristNoH}{0.0}

\newcommand{\MissWristSen}{16.1}

\newcommand{\MissWristThirtysecondH}{0.5}

\newcommand{\NEpisodes}{200}

\newcommand{\NSeeds}{8}

\newcommand{\NoHSenPMax}{0.034}
\newcommand{\NoiseCostHiMax}{3.6}

\newcommand{\NoiseRelMinDino}{+29.4}

\newcommand{\ObjSen}{\ensuremath{-}1.5}
\newcommand{\ObjSenHalf}{11.2}
\newcommand{\ObjSenHi}{+9.7}
\newcommand{\ObjSenLo}{\ensuremath{-}12.7}

\newcommand{\PickDinoCapacity}{+3.3}

\newcommand{\PickDinoFillRel}{+4.4}

\newcommand{\PickDinoFillValid}{+7.4}

\newcommand{\PickDinoFillValidP}{0.021}

\newcommand{\PickDinoGt}{+8.6}

\newcommand{\PickDinoGtP}{0.064}

\newcommand{\PickDinoGtRel}{+34.1}

\newcommand{\PickDinoGtSwapAnyRel}{+25.6}

\newcommand{\PickDinoInformation}{+5.2}
\newcommand{\PickDinoInformationP}{0.081}

\newcommand{\PickDinoNoFlagRel}{+22.9}

\newcommand{\PickDinoNoH}{+9.9}

\newcommand{\PickDinoNoHP}{0.008}

\newcommand{\PickDinoNoHRel}{+34.9}

\newcommand{\PickDinoNoHRgbBlind}{+24.9}

\newcommand{\PickDinoNoHSen}{+8.1}
\newcommand{\PickDinoNoHSenHi}{+11.3}
\newcommand{\PickDinoNoHSenLo}{+5.0}

\newcommand{\PickDinoNoHSwapAnyRel}{+29.8}

\newcommand{\PickDinoNoiseTwenty}{+7.3}

\newcommand{\PickDinoRgb}{76.2}

\newcommand{\PickDinoSen}{+1.8}

\newcommand{\PickDinoSenP}{0.547}

\newcommand{\PickDinoSenRel}{+2.7}

\newcommand{\PickDinoSenSwapAnyRel}{\ensuremath{-}0.4}

\newcommand{\PickMogeCapacity}{+2.8}

\newcommand{\PickMogeFillRel}{+23.2}

\newcommand{\PickMogeFillValid}{+10.6}

\newcommand{\PickMogeFillValidP}{0.01}

\newcommand{\PickMogeGt}{+8.8}

\newcommand{\PickMogeGtP}{0.064}

\newcommand{\PickMogeGtRel}{+52.8}

\newcommand{\PickMogeGtSwapAnyRel}{+30.4}

\newcommand{\PickMogeInformation}{+5.9}
\newcommand{\PickMogeInformationP}{0.153}

\newcommand{\PickMogeNoFlagRel}{+56.6}

\newcommand{\PickMogeNoH}{+10.0}

\newcommand{\PickMogeNoHP}{0.038}

\newcommand{\PickMogeNoHRel}{+51.8}

\newcommand{\PickMogeNoHRgbBlind}{+41.8}

\newcommand{\PickMogeNoHSen}{+9.7}
\newcommand{\PickMogeNoHSenHi}{+15.8}
\newcommand{\PickMogeNoHSenLo}{+3.5}

\newcommand{\PickMogeNoHSwapAnyRel}{+33.2}

\newcommand{\PickMogeNoiseTwenty}{+5.6}

\newcommand{\PickMogeRgb}{69.0}

\newcommand{\PickMogeSen}{+0.3}

\newcommand{\PickMogeSenP}{0.93}

\newcommand{\PickMogeSenRel}{+17.9}

\newcommand{\PickMogeSenSwapAnyRel}{+2.4}

\newcommand{\PickRnCapacity}{\ensuremath{-}7.1}
\newcommand{\PickRnCapacityP}{0.06}

\newcommand{\PickRnFillRel}{+5.3}

\newcommand{\PickRnFillValid}{+7.1}

\newcommand{\PickRnFillValidP}{0.03}

\newcommand{\PickRnGt}{+8.2}

\newcommand{\PickRnGtP}{0.078}

\newcommand{\PickRnGtRel}{+43.1}

\newcommand{\PickRnGtSwapAnyRel}{+9.2}

\newcommand{\PickRnInformation}{+15.3}
\newcommand{\PickRnInformationP}{0.003}
\newcommand{\PickRnNoFlag}{\ensuremath{-}5.1}

\newcommand{\PickRnNoFlagP}{0.054}
\newcommand{\PickRnNoFlagRel}{+16.0}

\newcommand{\PickRnNoH}{+9.9}

\newcommand{\PickRnNoHP}{0.005}

\newcommand{\PickRnNoHRel}{+40.4}

\newcommand{\PickRnNoHRgbBlind}{+30.5}

\newcommand{\PickRnNoHSen}{+11.2}
\newcommand{\PickRnNoHSenHi}{+19.8}
\newcommand{\PickRnNoHSenLo}{+2.7}

\newcommand{\PickRnNoHSwapAnyRel}{+8.1}

\newcommand{\PickRnNoiseTwenty}{+11.0}

\newcommand{\PickRnRgb}{47.9}

\newcommand{\PickRnSen}{\ensuremath{-}1.3}

\newcommand{\PickRnSenP}{0.757}

\newcommand{\PickRnSenRel}{+4.7}

\newcommand{\PickRnSenSwapAnyRel}{+1.9}

\newcommand{\RmseNoH}{0.22}

\newcommand{\RmseNoiseHundred}{22.37}

\newcommand{\RmseNoiseTwenty}{4.52}

\newcommand{\RmseSen}{0.25}

\newcommand{\SigmaGlobal}{2.86}
\newcommand{\SigmaRefWrist}{0.35}
\newcommand{\SigmaWrist}{0.19}

\newcommand{\StackRnCapacity}{+9.1}

\newcommand{\StackRnGt}{+17.1}

\newcommand{\StackRnGtN}{8}
\newcommand{\StackRnGtP}{0.004}

\newcommand{\StackRnGtRel}{+48.4}

\newcommand{\StackRnGtSwapAnyRel}{+13.8}

\newcommand{\StackRnInformation}{+7.9}

\newcommand{\StackRnNoH}{+19.3}

\newcommand{\StackRnNoHP}{0.008}

\newcommand{\StackRnNoHRel}{+47.6}

\newcommand{\StackRnNoHSen}{+12.9}
\newcommand{\StackRnNoHSenHi}{+24.4}
\newcommand{\StackRnNoHSenLo}{+1.3}

\newcommand{\StackRnNoHSwapAnyRel}{+14.5}

\newcommand{\StackRnRgb}{43.9}

\newcommand{\StackRnSen}{+6.4}

\newcommand{\StackRnSenP}{0.07}

\newcommand{\StackRnSenRel}{+2.8}

\newcommand{\StackRnSenSwapAnyRel}{+1.4}

\newcommand{\StructDinoPlace}{+0.5}

\newcommand{\StructMogePlace}{\ensuremath{-}1.0}

\newcommand{\StructRelMaxAbs}{2.6}

\newcommand{\StructRnPlace}{\ensuremath{-}2.2}
\newcommand{\StructRnPlaceHi}{+4.9}
\newcommand{\StructRnPlaceLo}{\ensuremath{-}9.4}

\newcommand{\WristHDinoVsNoH}{\ensuremath{-}6.2}
\newcommand{\WristHDinoVsNoHP}{0.022}

\newcommand{\WristHDinoVsSenP}{0.39}
\newcommand{\WristHMogeVsNoH}{\ensuremath{-}10.9}

\newcommand{\WristHRnVsNoH}{\ensuremath{-}13.2}

\newcommand{\WristPfifty}{14.1}

\newcommand{\XdBlindMin}{+10.0}

\newcommand{\XdDinoEighthH}{\ensuremath{-}17.1}

\newcommand{\XdDinoFourH}{\ensuremath{-}24.1}

\newcommand{\XdDinoSenRgb}{\ensuremath{-}14.2}
\newcommand{\XdDinoSenRgbP}{0.013}

\newcommand{\XdDinoThirtysecondH}{\ensuremath{-}6.4}

\newcommand{\XdMogeEighthH}{\ensuremath{-}28.4}

\newcommand{\XdMogeFourH}{\ensuremath{-}39.0}

\newcommand{\XdMogeSenRgb}{\ensuremath{-}27.1}
\newcommand{\XdMogeSenRgbP}{0.002}

\newcommand{\XdMogeThirtysecondH}{\ensuremath{-}10.1}

\newcommand{\XdRnEighthH}{\ensuremath{-}10.9}

\newcommand{\XdRnFourH}{\ensuremath{-}16.8}

\newcommand{\XdRnSenRgb}{\ensuremath{-}4.1}
\newcommand{\XdRnSenRgbP}{0.258}

\newcommand{\XdRnThirtysecondH}{\ensuremath{-}3.6}

\newcommand{\XdRnThirtysecondHP}{0.049}

\newcommand{\XnDinoGtNoiseTwenty}{+0.6}

\newcommand{\XnMogeGtNoiseTwenty}{\ensuremath{-}0.8}

\newcommand{\XnRnGtNoiseTwenty}{\ensuremath{-}1.1}

\newcommand{\RrEpisodes}{706}
\newcommand{\RrRate}{15}
\newcommand{\RrFamilies}{seven}

\newcommand{\RrSigmaNear}{0.17}
\newcommand{\RrSigmaNearD}{12.5}
\newcommand{\RrSigmaFar}{2.71}
\newcommand{\RrSigmaFarD}{62.5}

\newcommand{\RrHoleNear}{17.5}
\newcommand{\RrHoleFar}{3.4}
\newcommand{\RrHoleMax}{22.1}

\newcommand{\RrOpPfifty}{28.6}

\newcommand{\RrGrasp}{16.8}

\newcommand{\moge}{MoGe-2}

\makeatletter
\newcommand{\absval}[1]{\expandafter\absval@i#1\@nil}
\def\absval@i#1#2\@nil{\ifx+#1#2\else#1#2\fi}
\makeatother

\providecommand{\IEEEPARstart}[2]{#1#2}
\newenvironment{IEEEkeywords}{\setbox0\vbox\bgroup}{\egroup}

\title{\LARGE \bf Why Does Sensor Depth Fail Learned Manipulation Policies?\\
       Finding the Failure Mode with ACT}

\ifanonymous
  \author{Anonymous Authors}
\else
  \author{Inayat Rasool$^{1}$, Chulwoo Pack$^{2}$ and Shabbir Ahmed$^{1}$%
  \thanks{$^{1}$I. Rasool and S. Ahmed are with the Department of Mechanical
  Engineering, South Dakota State University, Brookings, SD 57007, USA.
  S. Ahmed ORCID: \href{https://orcid.org/0000-0001-8296-6025}{0000-0001-8296-6025}.}%
  \thanks{$^{2}$C. Pack is with the Department of Electrical Engineering and
  Computer Science, South Dakota State University, Brookings, SD 57007, USA.
  ORCID: \href{https://orcid.org/0000-0002-9876-4388}{0000-0002-9876-4388}.}%
  }
\fi

\begin{document}

\maketitle
\thispagestyle{empty}
\pagestyle{empty}

\begin{abstract}

Learned manipulation policies, from task-specific imitation policies to vision-language-action (VLA) models, rely primarily on RGB images. While some studies report gains from sensor depth, others avoid depth cameras because of measurement noise. Depth cameras introduce both noise and missing depth values (holes), whose relative effects remain unclear. In this study, we calibrate a sensor model using a real wrist-mounted stereo camera and separately evaluate noise and holes on RGB-D imitation policies with three visual encoders in simulation. We found that complete depth improves success by approximately ten percentage points, and most of this gain persists even at twenty times the camera's noise level. In contrast, holes covering about one-sixth of the wrist image eliminate the detectable benefit of depth. The validity mask strongly influences depth reliance: marking a small fraction of pixels as invalid reduces reliance, while filling holes and marking them valid restores most of the benefit. Policies trained on clean depth also lose performance when holes appear only at deployment, sometimes falling below RGB-only performance. These findings suggest that depth cameras for learned policies should therefore be judged by depth coverage more than measurement precision.
\end{abstract}

\begin{IEEEkeywords}
Perception for grasping and manipulation; RGB-D perception; deep learning for
visual perception; imitation learning.
\end{IEEEkeywords}

\section{Introduction}
\label{sec:intro}

\begin{figure*}[!t]
\centering
\includegraphics{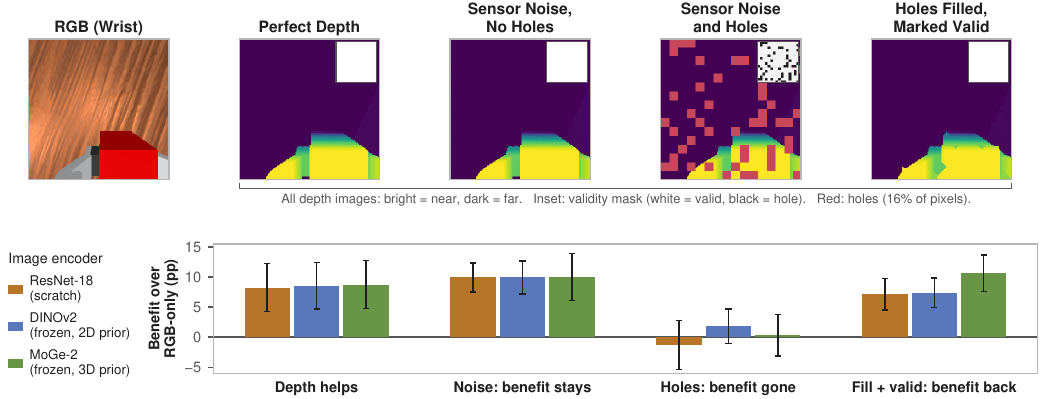}
\vspace{-0.3cm}
\caption{Effect of depth input quality on manipulation performance on \textsc{PickCube}. Top: Representative wrist-camera inputs, with missing depth values highlighted in red and validity masks shown in the insets (black indicates invalid pixels). Bottom: Change in task success rate relative to the RGB-only policy using the same encoder. Complete depth improves performance, whereas sensor-induced holes eliminate this benefit. Filling missing depth values and marking them as valid largely restores the improvement. Error bars represent one standard error across \NSeeds{} random seeds.}
\label{fig:teaser}
\end{figure*}

\IEEEPARstart{V}{ision}-language-action (VLA) models have emerged as a prominent approach
to robot manipulation and are pretrained on large-scale images and texts datasets, which gives them a strong sense of what objects are and what an instruction asks for \cite{rt2,openvla,pizero}. However, robotic manipulation is also a geometric problem, and a robot must know about the position of an object relative to its gripper to grasp or place it precisely. While an RGB image provides indirect cues about the pose and relative distance of a three-dimensional geometry, a depth sensor can directly measure distances in metric units. As a result, VLA models and task-specific manipulation policies are increasingly incorporating depth maps or point clouds as additional inputs.

However, existing studies differ on whether depth information should be acquired directly from sensors or inferred through learning-based policies. Studies such as 3D Diffusion Policy~\cite{ze2024dp3}, RISE~\cite{wang2024rise}, iDP3~\cite{ze2024idp3}, FP3~\cite{yang2025fp3}, 3D Diffuser Actor~\cite{ke2024dda}, Action Chunking with Transformers (ACT) and Vision Language Action (VLA) policies~\cite{pointact2025,chen2026pointact} etc. incorporates depth maps or point clouds through direct sensor measurements and report improved performance. In contrast, several other studies seek to eliminate the need for depth sensors during deployment by estimating or predicting depth from images, thereby distilling geometric information into the policy, or add 3D supervision during training for inferring depth \cite{depthvla2025,qdepthvla2025,peafowl2026,di22024,augvla2026,g3vla2026,spatialforcing2025,evodepth2026,stereopolicy2026}. A common motivation for these approaches is the susceptibility of depth sensors to measurement noise and missing depth information~\cite{evodepth2026,stereopolicy2026,qdepthvla2025,peafowl2026,spatialforcing2025}.



One possible explanation for these differing findings is that ``sensor quality'' encompasses two distinct limitations: measurement noise and missing depth values (holes). Measurement noise affects the accuracy of valid depth measurements, whereas holes occur when the sensor fails to return a depth value. A camera or sensor can be very precise where it provides a measured value, and at the same time, may leave a large part of the image without any depth information (holes).  Consequently, a sensor may provide accurate measurements while exhibiting substantial missing data. Previous studies have extensively characterized measurement accuracy of depth and associated noise~\cite{servi2021realsense,servi2024realsense,ahn2019d435},however, the impact of missing depth or holes on robotic manipulation remains less understood.

We therefore ask three questions: (1) Does complete depth information improve manipulation performance? (2) How do measurement noise and missing depth values (holes) independently affect this benefit? (3) What causes the performance degradation, and how can it be mitigated? To address these questions, we characterize the distance-dependent noise and missing-depth rate of a real wrist-mounted stereo camera on a teleoperated robot. Using noise-free depth rendered in simulation, we independently introduce measurement noise and missing values at the measured levels and under scaled perturbations. We evaluate Action Chunking with Transformers (ACT) policies~\cite{zhao2023act} using three visual encoders: ResNet-18 trained from scratch, DINOv2~\cite{oquab2024dinov2} with 2D pretraining, and \moge{}~\cite{moge2} with geometric pretraining. We quantify the contribution of depth using two metrics: \emph{benefit}, defined as the improvement in task success relative to an RGB-only policy with the same encoder, and \emph{reliance}, defined as the reduction in success when depth input is removed at deployment. All policy evaluations are conducted in simulation using ACT, without training VLA models.


Fig.~\ref{fig:teaser} summarizes the key findings. Complete depth information improves manipulation performance, and most of this benefit persists even when measurement noise increases to 20 times the measured level. In contrast, depth holes eliminate the benefit by causing policies to disregard depth information. This behavior is driven by the validity mask, which distinguishes missing depth from measured depth values. By filling the holes and marking them as valid restores most of the benefit. Moreover, policies trained on complete depth can perform worse than RGB-only policies when depth holes are introduced at deployment. These trends are consistent across all three visual encoders.


This paper makes four contributions: (1) a controlled evaluation of depth noise and holes, with a sensor model fitted to a real wrist camera; (2) provide evidence that the holes, not noise, remove the benefit of sensor depth for the studied policies; (3) provide evidence that validity masks cause policies to disregard incomplete depth information; and (4) provide practical guidance for depth sensor selection and preprocessing, along with a metric (\emph{reliance}) to quantify whether a trained policy uses depth information.
\section{Study Design}
\label{sec:design}

\begin{figure*}[!t]
\centering
\includegraphics[width=\textwidth]{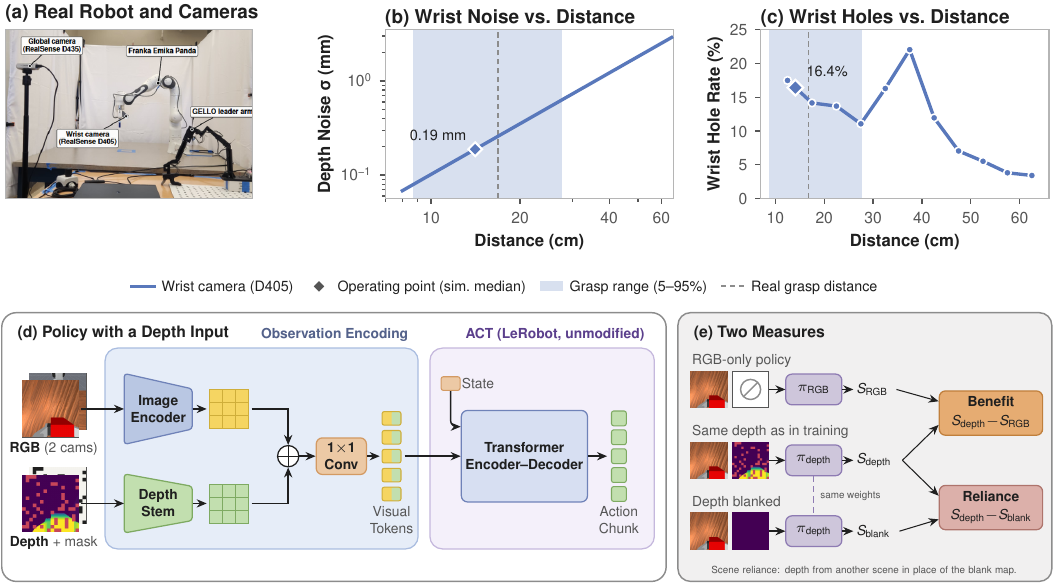}
\caption{Real camera, sensor model and policy. \textbf{(a)} Setup for the
real-robot measurements. A GELLO leader arm teleoperates a Franka Emika Panda,
with a D405 at the wrist and a D435 on a fixed mount. \textbf{(b)} Wrist depth
noise against distance, from the fitted model. \textbf{(c)} Wrist hole rate per
5\,cm bin. In (b) and (c), the band is the grasp range in simulation, and the
dashed line is the real grasp distance. \textbf{(d)} Depth and its validity
mask go through a small depth stem, whose features are joined to the image
features of an unmodified ACT policy. \textbf{(e)} Benefit compares a depth
policy with an RGB-only policy. Reliance compares the same policy with its
depth intact and with its depth blanked or taken from another scene.}
\label{fig:design}
\end{figure*}

To isolate the effects of individual depth input qualities, experiments need to be designed such that one property varies at a time while holding all others constant. Since measurement noise and missing depth values cannot be independently controlled on a physical camera, we render ground-truth depth in the ManiSkill3 simulator~\cite{maniskill3} and introduce controlled degradations using a model calibrated to real camera measurements. The experimental setup consists of a 7 degrees-of-freedom Franka Emika Panda robotic arm equipped with a wrist-mounted camera and a third-person camera. We employ a factorial design across depth conditions and visual encoders, maintaining identical policy architectures, demonstrations, action spaces, evaluation episodes, and training schedules across all configurations.


\subsection{Sensor Model Fitted to a Real Wrist Camera}
\label{sec:sensor}

We recorded \RrEpisodes{} teleoperated episodes at \RrRate~Hz on a Franka
Emika Panda driven by a GELLO leader arm~\cite{wu2024gello}
(Fig.~\ref{fig:design}a). The episodes cover \RrFamilies{} task families,
such as pick-and-place, stacking, pushing and drawer opening. The wrist
camera is an Intel RealSense D405, a short-range stereo camera with an 18\,mm
baseline and a rated range of 7--50\,cm. The third-person camera is a
RealSense D435.

To quantify depth measurement noise, we select a fixed $6\times6$,cm$^2$ patch on the table and record its depth measurements across successive video frames. For each frame, a best-fit plane is estimated from the valid 3D points within the patch, and the noise level $\sigma$ is defined as the root-mean-square distance of these points from the fitted plane. As we move the wrist camera every 5 cm distance, $\sigma$ increases from
\RrSigmaNear\,mm at \RrSigmaNearD\,cm to \RrSigmaFar\,mm at \RrSigmaFarD\,cm. This trend is approximated by the power law $\sigma(d) = \SigmaRefWrist,(d/20,\mathrm{cm})^{\KWrist}$mm (Fig.~\ref{fig:design}b), with an exponent close to 2, consistent with the expected quadratic dependence of stereo-triangulation depth uncertainty on distance.




The depth hole rate is defined as the fraction of pixels in each frame without valid depth measurements (Fig.~\ref{fig:design}c). For the wrist camera, the hole rate ranges from \RrHoleNear\% in the nearest distance to \RrHoleFar\% in the farthest, reaching a maximum of \RrHoleMax\% at intermediate distances. The rate also depends on surface characteristics, as stereo matching relies on visual texture and can fail on reflective or transparent surfaces. At the simulated robot's median working distance, the calibrated model predicts a depth noise level of \SigmaWrist\,mm and a hole rate of \HoleWrist\% for the wrist camera, compared with \SigmaGlobal\,mm and \HoleGlobal\%, respectively, for the third-person camera. Thus, the wrist camera provides approximately 15 times lower depth noise but exhibits about 8 times higher proportion of holes.

Using the calibrated sensor model, we independently introduce depth noise and missing values into the simulated depth maps. Measurement noise is modeled as zero-mean Gaussian noise applied independently to each pixel, with the standard deviation $\sigma$ determined by its depth. This model excludes systematic bias, spatial noise correlation, and artifacts near depth discontinuities, such as flying pixels. Missing depth values are simulated by partitioning the $128\times128$ policy input into nonoverlapping $8\times8$ pixel blocks, each independently designated as invalid with a probability equal to the frame's prescribed hole rate. The hole pattern is resampled for each frame, independently of scene content.


The simulated wrist camera operates at shorter distances than the real camera, with median distances of \WristPfifty cm and \RrOpPfifty cm, respectively. However, the simulated distances overlap with the real camera's near-field operating range, including grasping at approximately \RrGrasp cm. Since noise and hole rates are modeled as functions of distance, the simulated depth degradation reflects the measured sensor characteristics at corresponding distances. Even at the farthest real operating distance, the noise remains within the evaluated noise range, while the hole rate is approximately 6\%.


\subsection{Depth Conditions}
\label{sec:conditions}

\begin{table}[t]
\caption{Depth input conditions. All conditions except RGB-only give the policy depth and a
validity mask for each camera.}
\label{tab:conditions}
\centering
\footnotesize
\setlength{\tabcolsep}{3pt}
\begin{tabular}{@{}>{\raggedright\arraybackslash}p{0.30\columnwidth}>{\raggedright\arraybackslash}p{0.49\columnwidth}l@{}}
\toprule
Condition & What the policy receives & Used in \\
\midrule
RGB-only & Color only. Baseline for each encoder. & all \\
Perfect depth & Exact simulator depth. The ceiling. & III-A \\
Constant depth & One fixed value per camera. Depth pathway, no information. & III-A \\
Hole-free sensor depth & Exact depth with the fitted noise. No holes. & III-A--E \\
Sensor depth & Fitted noise and holes at the measured rate. & III-B--E \\
Noise $\times k$ & Fitted noise $\times$5, 20, 50, 100. No holes. & III-B \\
Holes $\times k$ & Hole rate $\times$1/8 to 4 (also 1/16, 1/32 for ResNet-18). Fitted noise. & III-B, C, E \\
Holes at wrist only & Sensor depth; no holes at the third-person camera. & III-B \\
Filled, mask correct & Holes filled from nearest valid pixel; mask marks them. & III-D \\
Empty, marked valid & Holes left empty; mask marks every pixel valid. & III-D \\
Filled, marked valid & Holes filled; mask marks every pixel valid. & III-D \\
\bottomrule
\end{tabular}
\end{table}



Table~\ref{tab:conditions} summarizes the depth conditions used in this study. To investigate the effect of missing depth values (holes), we compare two conditions: sensor depth, which contains both noise and holes based on the calibrated camera model, and hole-free sensor depth, which contains the same noise but no holes. During training, the noise is generated using a fixed seed for each frame, ensuring identical noise values in both conditions. Across evaluation rollouts, the wrist-camera depth RMSE is \RmseNoH mm without holes and \RmseSen mm with holes, while the corresponding hole rates are \MissWristNoH\% and \MissWristSen\%, respectively.


Two additional experiments independently vary noise and hole severity. The noise sweep scales measurement noise by factors of 5, 20, 50, and 100 without introducing holes, reaching a wrist-camera RMSE of \RmseNoiseHundred mm. The hole sweep scales the calibrated hole rate from 1/8 to 4, producing missing-depth rates of \MissWristEighthH\% to \MissWristFourH\%. Additional levels of 1/16 and 1/32 are used for ResNet-18, while all encoders are evaluated across all levels at deployment (Sec.~\ref{sec:mismatch}). Perfect depth uses the simulator's ground-truth measurements, whereas constant depth assigns a fixed value to each camera to retain the depth input pathway without providing geometric information.

\subsection{Policy and Depth Input}
\label{sec:policy}

We use the ACT policy~\cite{zhao2023act} implemented in LeRobot~\cite{lerobot}, modifying only the visual feature extraction module (Fig.~\ref{fig:design}d). Each camera provides two depth channels: metric depth, normalized over fixed ranges of 0--2 m for the wrist camera and 0--10 m for the third-person camera, and a binary validity mask indicating missing measurements. Fixed normalization preserves the magnitude of depth errors, while the validity mask distinguishes missing values from valid measurements. A lightweight trainable depth encoder comprising three convolutional blocks (approximately 0.2,M parameters) extracts depth features, which are concatenated with RGB features before the policy projection layer, allowing each visual token to incorporate both color and depth information.

To evaluate the influence of visual pretraining, we consider three image encoders: ResNet-18 trained from scratch, DINOv2~\cite{oquab2024dinov2}, a frozen ViT-B/14 pretrained on 2D imagery, and \moge{}~\cite{moge2}, a frozen ViT-B/14 further trained from DINOv2 weights for geometric understanding. Comparing DINOv2 and \moge{} isolates the effect of geometry-oriented pretraining, while ResNet-18 provides a baseline without pretraining.

\subsection{Tasks, Training and Evaluation}
\label{sec:tasks}


\textsc{StackCube}, which requires stacking one cube on another, is evaluated using ResNet-18 across the main depth conditions to assess generalization to a second task. Motion-planning demonstrations are used for both tasks, with \DemosPick{} for \textsc{PickCube} and \DemosStack{} for \textsc{StackCube}, selected to maintain moderate RGB-only success rates. Images from both cameras are rendered at $256\times256$ pixels and resized to $128\times128$ before depth corruption. The policy predicts 16-action chunks, executes 8 actions per cycle, and controls changes in end-effector pose. Training uses AdamW for 30{,}000 steps with a batch size of 64 and learning rates of $10^{-4}$, or $10^{-5}$ for trainable visual encoders and the depth stem. The checkpoint with the lowest validation loss is selected using 10\% held-out demonstrations. Each configuration is trained with \NSeeds{} random seeds and evaluated over \NEpisodes{} episodes per seed using identical scenes. An episode is considered successful if the task's success criterion is met before the step limit.

\subsection{Measures and Statistics}
\label{sec:measures}

Let $S_c$ denote the success rate (\%) of a policy trained under condition $c$, $S_c^{\varnothing}$ its success when depth is replaced by a blank map, and $S_{\mathrm{RGB}}$ the success of the RGB-only policy with the same encoder and seed. We define \emph{benefit} and \emph{reliance}, in percentage points (pp), as
\begin{equation}
B_c = \mathbb{E}s\bigl[S_c - S{\mathrm{RGB}}\bigr], \qquad
R_c = \mathbb{E}_s\bigl[S_c - S_c^{\varnothing}\bigr],
\label{eq:measures}
\end{equation}
where $\mathbb{E}_s$ denotes the mean over \NSeeds{} training seeds. Benefit measures improvement over RGB-only performance, including any effect of the additional depth-processing parameters, whereas reliance measures sensitivity to depth removal using the same trained policy and evaluation episodes. The blank map assigns the camera's median depth for that frame to all pixels, marked as valid.

To reduce the influence of out-of-distribution blank inputs, we also evaluate \emph{scene reliance} by replacing depth with a corrupted depth map from a randomly selected step in another demonstration scene. The \emph{halving point} is the hole rate at which reliance falls to half its hole-free value, estimated by linear interpolation with bootstrap confidence intervals over seeds.

Statistical comparisons use training runs as independent replicates. For each seed, success differences are computed on matched evaluation episodes and assessed using a one-sample $t$-test across \NSeeds{} seeds, with Holm correction for multiple comparisons addressing the same question. This avoids treating evaluation episodes from the same trained policy as independent training replicates.
\section{Results and Discussions}
\label{sec:results}

 \subsection{Complete Depth Helps}
 \label{sec:benefit}


\begin{table}[t]
\caption{Benefit and reliance (percentage points, pp), with $p$-values from paired $t$-tests across \NSeeds{} training seeds. The contrast row reports the difference between hole-free and sensor-depth conditions. All four contrasts remain statistically significant after Holm correction for multiple comparisons.}
\label{tab:main}
\centering
\footnotesize
\setlength{\tabcolsep}{1.5pt}
\begin{tabular}{@{}lcccc@{}}
\toprule
 & \multicolumn{3}{c}{PickCube} & Stack \\
\cmidrule(lr){2-4}\cmidrule(l){5-5}
 & ResNet & DINOv2 & MoGe-2 & ResNet \\
\midrule
RGB-only success, \% & \PickRnRgb & \PickDinoRgb & \PickMogeRgb & \StackRnRgb \\
\midrule
\multicolumn{5}{@{}l}{\emph{Benefit over RGB-only}} \\
Perfect & \PickRnGt & \PickDinoGt & \PickMogeGt & \StackRnGt \\
\quad $p$ & $\PickRnGtP$ & $\PickDinoGtP$ & $\PickMogeGtP$ & $\StackRnGtP$ \\
Hole-free sensor & \PickRnNoH & \PickDinoNoH & \PickMogeNoH & \StackRnNoH \\
\quad $p$ & $\PickRnNoHP$ & $\PickDinoNoHP$ & $\PickMogeNoHP$ & $\StackRnNoHP$ \\
Sensor & \PickRnSen & \PickDinoSen & \PickMogeSen & \StackRnSen \\
\quad $p$ & $\PickRnSenP$ & $\PickDinoSenP$ & $\PickMogeSenP$ & $\StackRnSenP$ \\
\textbf{Hole-free $-$ sensor} & \textbf{\PickRnNoHSen} & \textbf{\PickDinoNoHSen} & \textbf{\PickMogeNoHSen} & \textbf{\StackRnNoHSen} \\
\quad\scriptsize 95\% CI & {\tiny[\PickRnNoHSenLo, \PickRnNoHSenHi]} & {\tiny[\PickDinoNoHSenLo, \PickDinoNoHSenHi]} & {\tiny[\PickMogeNoHSenLo, \PickMogeNoHSenHi]} & {\tiny[\StackRnNoHSenLo, \StackRnNoHSenHi]} \\
\midrule
\multicolumn{5}{@{}l}{\emph{Reliance (depth blanked)}} \\
Perfect & \PickRnGtRel & \PickDinoGtRel & \PickMogeGtRel & \StackRnGtRel \\
Hole-free sensor & \PickRnNoHRel & \PickDinoNoHRel & \PickMogeNoHRel & \StackRnNoHRel \\
Sensor & \PickRnSenRel & \PickDinoSenRel & \PickMogeSenRel & \StackRnSenRel \\
\midrule
\multicolumn{5}{@{}l}{\emph{Scene reliance (another scene's depth)}} \\
Perfect & \PickRnGtSwapAnyRel & \PickDinoGtSwapAnyRel & \PickMogeGtSwapAnyRel & \StackRnGtSwapAnyRel \\
Hole-free sensor & \PickRnNoHSwapAnyRel & \PickDinoNoHSwapAnyRel & \PickMogeNoHSwapAnyRel & \StackRnNoHSwapAnyRel \\
Sensor & \PickRnSenSwapAnyRel & \PickDinoSenSwapAnyRel & \PickMogeSenSwapAnyRel & \StackRnSenSwapAnyRel \\
\bottomrule
\end{tabular}
\end{table}

We first evaluate the benefit of complete depth information (Table~\ref{tab:main}). Perfect depth improves mean success across all policies, with gains of \absval{\PickRnGt}, \absval{\PickDinoGt}, and \absval{\PickMogeGt} pp for ResNet-18, DINOv2, and \moge{} on \textsc{PickCube}, respectively, although none is individually significant ($p=\PickDinoGtP$ to $\PickRnGtP$). On \textsc{StackCube}, the gain reaches \absval{\StackRnGt} pp ($p=\StackRnGtP$) is consistent across all \StackRnGtN{} seeds, and remains significant after Holm correction. Hole-free sensor depth yields comparable improvements of \absval{\PickRnNoH}--\absval{\PickMogeNoH} pp on \textsc{PickCube} significant for each encoder before correction, and \absval{\StackRnNoH} pp on \textsc{StackCube}.

To distinguish the contribution of geometric information from that of additional network capacity, we compare perfect depth with constant depth, which preserves the depth-processing pathway without providing scene geometry. Perfect depth improves success by \absval{\PickRnInformation}, \absval{\PickDinoInformation}, and \absval{\PickMogeInformation} pp on \textsc{PickCube}, and \absval{\StackRnInformation} pp on \textsc{StackCube}. Only the ResNet-18 comparison on \textsc{PickCube} is significant ($p=\PickRnInformationP$); the other \textsc{PickCube} comparisons yield $p=\PickDinoInformationP$ to $\PickMogeInformationP$. Relative to RGB-only, constant depth changes success by \PickRnCapacity, \PickDinoCapacity, and \PickMogeCapacity pp on \textsc{PickCube}, and \StackRnCapacity pp on \textsc{StackCube}, indicating that the depth pathway itself can affect performance. In particular, the unexplained performance reduction for ResNet-18 with constant depth ($p=\PickRnCapacityP$) makes its estimated information gain larger than its overall benefit. On \textsc{StackCube}, approximately half the observed benefit is associated with the additional pathway rather than geometric information. Overall, the results suggest depth-related improvements of approximately 5--15 pp, consistent with prior studies~\cite{zhu2024pcm,ze2024dp3,wang2024rise}. Stronger evidence for the contribution of depth information comes from the controlled hole comparisons (Sec.~\ref{sec:holes}) and reliance analysis.

\subsection{Holes, Not Noise, Remove the Benefit}
\label{sec:holes}

\begin{figure*}[!t]
\centering
\includegraphics{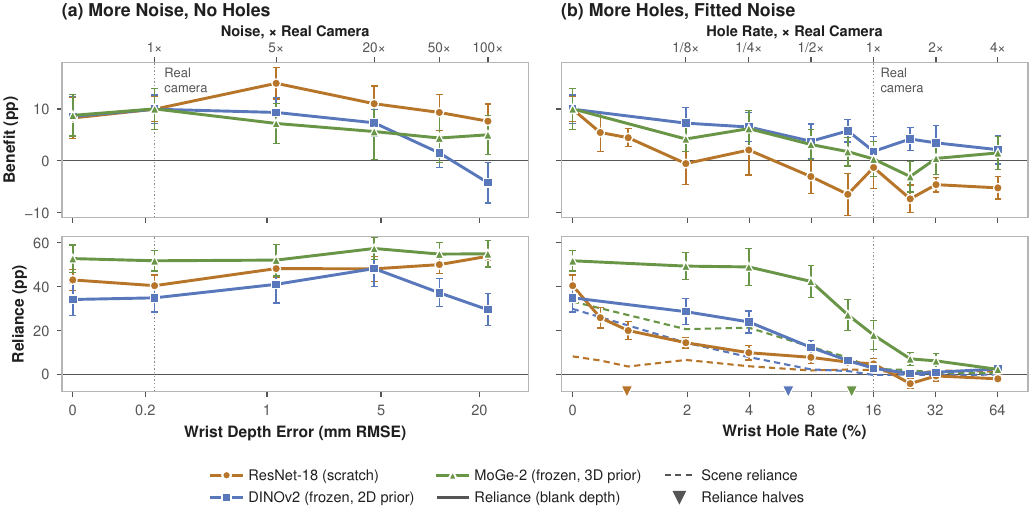}
\vspace{-0.5cm}
\caption{Effects of depth noise and missing values on \textsc{PickCube} performance. \textbf{(a)} Noise scaled by factors of 5--100 without holes. \textbf{(b)} Hole rates scaled up to four times the calibrated level at fixed noise. The top panels show the benefit relative to RGB-only policies, while the bottom panels show reliance on depth removal (solid lines) and, in (b), scene reliance (dashed lines). Triangles indicate the halving points of blank-depth reliance, and the dotted line marks the calibrated camera level. Noise largely preserves the benefit and reliance on depth, whereas holes substantially reduce both. Error bars represent one standard error across \NSeeds{} training seeds.}
\label{fig:sweeps}
\end{figure*}

It was found that the measurement noise has little effect on manipulation performance. Hole-free sensor depth performs comparably to perfect depth across all encoders and tasks, with no significant differences. Even at 20 times the calibrated noise level (wrist RMSE of \RmseNoiseTwenty mm), depth provides benefits of \absval{\PickRnNoiseTwenty}, \absval{\PickDinoNoiseTwenty}, and \absval{\PickMogeNoiseTwenty} pp for ResNet-18, DINOv2, and \moge{}, respectively, with statistically significant improvements for two encoders before correction for multiple comparisons. At 100 times the noise (\RmseNoiseHundred mm), none of the benefits remain significant, although policies continue to use depth, with reliance at least \absval{\NoiseRelMinDino} pp throughout the sweep (Fig.~\ref{fig:sweeps}a). Policies trained on perfect depth also show limited sensitivity to noise introduced only at deployment, with success changes of \XnRnGtNoiseTwenty, \XnDinoGtNoiseTwenty, and \XnMogeGtNoiseTwenty pp at 20 times the noise. These findings indicate that depth precision is not the primary limitation within the tested range, although the noise model excludes spatially correlated and structured errors.

In contrast, missing depth values substantially reduce performance. At the calibrated wrist-camera hole rate of \MissWristSen\%, sensor-depth policies achieve improvements over RGB-only policies of \PickRnSen, \PickDinoSen, and \PickMogeSen pp on \textsc{PickCube}, and \StackRnSen pp on \textsc{StackCube}, none of which is statistically significant. However, a direct comparison with hole-free sensor depth reveals performance reductions of \absval{\PickRnNoHSen}, \absval{\PickDinoNoHSen}, and \absval{\PickMogeNoHSen} pp on \textsc{PickCube}, and \absval{\StackRnNoHSen} pp on \textsc{StackCube}, demonstrating the adverse effect of missing depth information. All four differences remain significant after Holm correction ($p\leq\NoHSenPMax$), although their magnitudes vary across seeds. To identify the source of this degradation, we introduce holes only in the wrist-camera depth. Performance is statistically indistinguishable from introducing holes in both cameras ($p\geq\WristHDinoVsSenP$), while comparisons with hole-free depth yield changes of \WristHRnVsNoH, \WristHDinoVsNoH, and \WristHMogeVsNoH pp ($p\leq\WristHDinoVsNoHP$). This indicates that the observed degradation is primarily associated with missing wrist-camera depth, despite its higher measurement precision. Whether the resulting loss reflects misleading depth information or reduced reliance on depth is examined next.

Replacing depth with a blank map can overestimate a policy's reliance on depth information, as the unfamiliar input may itself degrade performance. For policies trained on hole-free sensor depth, blanking reduces success below the RGB-only baseline by \absval{\PickRnNoHRgbBlind}, \absval{\PickDinoNoHRgbBlind}, and \absval{\PickMogeNoHRgbBlind} pp on \textsc{PickCube}, suggesting that approximately three-quarters of the measured reliance may reflect sensitivity to out-of-distribution inputs. We therefore emphasize \emph{scene reliance}, which replaces depth with measurements from another scene while preserving realistic input characteristics. For hole-free sensor depth, scene reliance is \absval{\PickRnNoHSwapAnyRel}, \absval{\PickDinoNoHSwapAnyRel}, and \absval{\PickMogeNoHSwapAnyRel},pp on \textsc{PickCube}, and \absval{\StackRnNoHSwapAnyRel},pp on \textsc{StackCube}, all statistically significant. ResNet-18 exhibits substantially lower scene reliance than the pretrained encoders, indicating weaker use of depth geometry. In contrast, policies trained with sensor-depth holes exhibit scene reliance of only \PickRnSenSwapAnyRel, \PickDinoSenSwapAnyRel, and \PickMogeSenSwapAnyRel pp on \textsc{PickCube}, and \StackRnSenSwapAnyRel pp on \textsc{StackCube}. These results suggest that policies trained with missing depth rely primarily on RGB information rather than exploiting the available depth measurements. Consistently, across 69 depth-trained configurations, 17 show significant benefits ($p<0.05$), of which only two exhibit blank-depth reliance below 10 pp.

The hole sweep further quantifies how reliance decreases with missing depth (Fig.~\ref{fig:sweeps}b). Scene reliance falls to half its hole-free value at hole rates of \HalfRelSwapAnyRn\%, \HalfRelSwapAnyDino\%, and \HalfRelSwapAnyMoge\% for ResNet-18, DINOv2, and \moge{}, respectively, with corresponding 95\% bootstrap intervals of [\HalfRelSwapAnyRnLo, \HalfRelSwapAnyRnHi], [\HalfRelSwapAnyDinoLo, \HalfRelSwapAnyDinoHi], and [\HalfRelSwapAnyMogeLo, \HalfRelSwapAnyMogeHi]\%. Blank-depth reliance reaches the same threshold at higher hole rates: \HalfRelRn\%, \HalfRelDino\%, and \HalfRelMoge\%, respectively. On \textsc{StackCube}, the scene-reliance and blank-depth halving points are \HalfRelSwapAnyStackRn\% and \HalfRelStackRn\%. All point estimates fall below the calibrated wrist-camera hole rate of \MissWristSen\%, indicating that the measured level of missing depth is sufficient to substantially reduce depth reliance. These thresholds are specific to the simulated random-hole model, and their uncertainty is limited by the \NSeeds{} training seeds. The findings motivate a closer examination of which depth-input components cause policies to disregard geometric information.

\subsection{Effect of the Validity Mask}
\label{sec:mask}

\begin{figure*}[!t]
\centering
\includegraphics{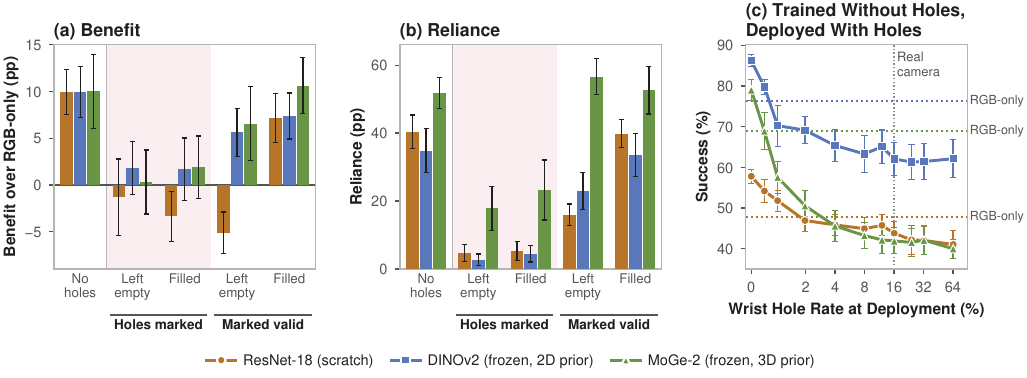}
\vspace{-0.5cm}
\caption{Effect of the validity mask, and of holes that appear only at
deployment on \textsc{PickCube}. \textbf{(a)} Benefit over RGB-only policies;
\textbf{(b)} Reliance with blank depth. Holes are left empty or filled from the
nearest valid pixel, and the mask either marks them (shaded) or marks every
pixel as valid. Hole-free sensor depth is the reference. \textbf{(c)} Success
of policies trained without holes and deployed with them. Dotted lines show the
RGB-only policy. Marking every pixel as valid raises reliance, but only filling
the holes as well restores the benefit. Error bars show one standard error over
\NSeeds{} seeds.}
\label{fig:mask}
\end{figure*}

A hole reaches the policy as a zero in the depth values and as a mark in the
validity mask. We separate the two with four conditions on the same sensor
depth (Table~\ref{tab:conditions}, Fig.~\ref{fig:mask}a,b). The hole is left
empty or filled from the nearest valid pixel, and the mask is correct or marks
every pixel valid.

The mask has the larger effect on reliance. Averaged over the two hole values,
a mask that marks every pixel valid increases reliance by
\absval{\FlagRnValid}, \absval{\FlagDinoValid} and \absval{\FlagMogeValid}\,pp
for ResNet-18, DINOv2 and \moge{} ($p \le \FlagDinoValidP$). Filled values
increase it by \absval{\FlagRnValue}\,pp for ResNet-18 ($p = \FlagRnValueP$),
and by \absval{\FlagDinoValue} and \absval{\FlagMogeValue}\,pp for the
pretrained encoders. The two act together, and reliance returns to its
hole-free level only when the holes are both filled and marked valid.

Filled holes with a correct mask do not help. Their reliance is
\absval{\PickRnFillRel}, \absval{\PickDinoFillRel} and
\absval{\PickMogeFillRel}\,pp, against \absval{\PickRnNoHRel},
\absval{\PickDinoNoHRel} and \absval{\PickMogeNoHRel}\,pp for depth that never
had holes, and no benefit differs measurably from zero. The policy ignores
filled depth that the mask marks, as it ignores depth with holes. Empty holes that are marked valid make the
policy use depth, and reliance increases to \absval{\PickRnNoFlagRel},
\absval{\PickDinoNoFlagRel} and \absval{\PickMogeNoFlagRel}\,pp. However, no
benefit is significant, and ResNet-18 changes by \PickRnNoFlag\,pp
($p = \PickRnNoFlagP$), a possible cost, because the depth now says that a
sixth of the wrist image is a surface at zero distance. Filled holes that are
marked valid recover the benefit, with gains of \absval{\PickRnFillValid}\,pp
($p = \PickRnFillValidP$), \absval{\PickDinoFillValid}\,pp
($p = \PickDinoFillValidP$) and \absval{\PickMogeFillValid}\,pp
($p = \PickMogeFillValidP$), significant before correction. These are close to the \absval{\PickRnNoH},
\absval{\PickDinoNoH} and \absval{\PickMogeNoH}\,pp of hole-free sensor depth,
and the reliance is also close to the hole-free values.

Thus a policy stops using depth when the mask marks part of the image as
missing, and it benefits from depth when every pixel is marked valid and the
values are plausible. Multimodal networks are known to ignore one of their
inputs~\cite{wu2022greedy,liu2025factr,fei2025liberoplus}. Our result gives a
cause for depth, which is a mark that tells the network in every frame that
the input is incomplete. The result holds for random hole placement, and a
fill on hardware can put invented depth on the object to grasp
(Sec.~\ref{sec:limits}).

\subsection{A Policy Keeps the Decision It Made in Training}
\label{sec:mismatch}

In the above analysis, each policy has the same holes in training and at deployment. A policy might also be trained on complete depth, from simulation or from cleaned data, and then deployed on a real sensor. We therefore deploy each policy trained on hole-free sensor depth at every level of the hole sweep (Fig.~\ref{fig:mask}c). With \MissWristThirtysecondH\% holes, success changes
by \XdRnThirtysecondH, \XdDinoThirtysecondH{} and \XdMogeThirtysecondH\,pp for ResNet-18, DINOv2 and \moge{} ($p \le \XdRnThirtysecondHP$). With \MissWristEighthH\% holes the change is \XdRnEighthH, \XdDinoEighthH{} and \XdMogeEighthH\,pp, and with \MissWristFourH\% it is \XdRnFourH, \XdDinoFourH{} and \XdMogeFourH\,pp. These policies stay better than the same weights with blank depth by at least \absval{\XdBlindMin}\,pp, so they still use the depth that is left.

The important comparison is with a policy that never had depth. At the camera's hole rate, success changes against the RGB-only policy by \XdMogeSenRgb\,pp for \moge{} ($p = \XdMogeSenRgbP$) and by
\XdDinoSenRgb\,pp for DINOv2 ($p = \XdDinoSenRgbP$), both significant before
correction, while ResNet-18 changes by \XdRnSenRgb\,pp ($p = \XdRnSenRgbP$). A policy trained on clean depth and deployed on a real sensor can thus be worse than a policy with no depth. The pretrained encoders, which take the most scene information from depth, lose the most. The reverse case gives no gain. We deploy each policy trained with holes on depth with the holes removed. In \GainN{} such cases, no change is larger than \GainAbsMax\,pp, and none is significant after a Holm correction.

These results suggest that, during training, the policy learns how much to rely on depth and largely retains this behavior at deployment. Depth noise has little effect on this reliance, whereas a validity mask indicating missing regions in the wrist image substantially reduces it. This interpretation is consistent with the observed results, although we did not examine the network's internal representations to verify the underlying mechanism.

\subsection{Outlook for VLA Models}
\label{sec:vla}


Our experiments use ACT rather than a VLA model, so the implications discussed here remain hypotheses that require direct validation. Nevertheless, several architectural components relevant to our findings are shared with existing VLA models. The frozen DINOv2 encoder is one of the two visual encoders used in OpenVLA~\cite{openvla}, while recent VLA fine-tuning approaches adopt ACT's chunked, regression-based action head~\cite{kim2025oft}. When depth is introduced into a pretrained VLA during fine-tuning, the new depth pathway must learn to contribute alongside an RGB pathway that already provides substantial task-relevant information. Based on our findings, we therefore expect that a VLA that receives depth as image tokens will learn to ignore these tokens if its wrist camera marks a sixth of the view as invalid, as the camera in this study does.


\subsection{Limitations}
\label{sec:limits}





Although the depth corruption model is calibrated using measurements from a real camera, all policy evaluations are conducted in simulation, with demonstrations generated by a motion planner. The corruption model also simplifies several characteristics of real depth measurements. Specifically, holes are represented as square blocks, placed independently of the scene and resampled at every frame. In contrast, real depth holes often form larger connected regions, occur more frequently on flat or nearby surfaces, and persist on the same surfaces over time. To assess the effect of hole placement, we conducted an additional experiment in which holes followed the spatial pattern observed in the real camera, while keeping the overall hole rate unchanged. Compared with random placement, success changed by \StructRnPlace, \StructDinoPlace{}, and \StructMogePlace,pp for ResNet-18, DINOv2, and \moge{}, respectively, with none of these differences statistically significant. Reliance changed by at most \StructRelMaxAbs,pp. However, these findings do not rule out an effect of hole placement. The confidence interval for ResNet-18, [\StructRnPlaceLo, \StructRnPlaceHi],pp, includes a possible performance loss. Furthermore, the comparison was performed at the camera's observed hole rate, where the policies already showed little reliance on depth regardless of hole placement. Therefore, the estimated halving points and the observed benefits of hole filling remain specific to random hole placement. Holes concentrated on the object being grasped could lead to different outcomes.

The noise model introduces independent errors at each pixel and therefore does not capture spatially or temporally correlated depth errors. It also does not account for the camera's minimum sensing distance of approximately 7,cm, below which valid depth measurements are unavailable. Beyond these modeling assumptions, the study is limited to a single policy class, one camera model, and two grasping tasks. We additionally evaluated a third task involving planar pushing, but the policies showed neither measurable reliance on depth nor a performance benefit from it across three seeds. These limitations motivate further evaluation with real robotic hardware, more realistic depth corruption patterns, and a broader range of manipulation tasks.
\section{Conclusions}
\label{sec:conclusion}


This study reconciles two competing perspectives on the value of depth sensing in learned robotic manipulation. While complete depth information improves policy performance, depth from a real wrist-mounted camera may provide little or no benefit. By separating sensor noise from missing depth measurements, we show that the loss of benefit is primarily associated with depth holes rather than measurement noise. Even when noise is increased to twenty times the camera's measured level, most of the benefit remains, with a performance cost bounded by \NoiseCostHiMax pp (one-sided 95\% bound). In contrast, depth holes substantially reduce the policy's reliance on depth, particularly when the validity mask identifies missing regions. Filling these holes and marking the corresponding pixels as valid restores most of the benefit. These findings suggest that sensor noise alone is not a compelling reason to avoid depth sensing in the setting studied here, although we do not directly compare sensor depth with alternative depth-estimation approaches.

The overall trends are consistent across all three visual encoders, although their sensitivity to missing depth differs. The estimated halving points range from \HalfRelSwapAnyRn\% to \HalfRelMoge\%, while the pretrained encoders extract more task-relevant information from depth and experience larger performance losses when holes appear only at deployment. Whether an explicit 3D prior improves the usefulness of sensor depth remains unresolved. The measured difference between DINOv2 and \moge{} is \ObjSen pp, with a 95\% confidence interval of [\ObjSenLo, \ObjSenHi] pp, which is too wide to establish or rule out an effect. Detecting differences smaller than approximately \ObjSenHalf pp would require additional seeds.

These results have several practical implications for integrating depth sensors into learned manipulation policies. First, camera selection should prioritize depth coverage at the robot's working distance rather than measurement precision alone, since the fraction of missing depth can be more consequential than sensor noise. Second, missing depth values should be filled before being passed to the policy, with the corresponding pixels marked as valid. Although the original validity mask remains necessary to identify holes during preprocessing, retaining these invalid markings in the policy input can suppress depth utilization even after filling. Third, training and deployment should use consistent depth preprocessing, including realistic missing-depth patterns and the same hole-filling procedure, to reduce the mismatch between simulated or recorded training data and real sensor measurements. Finally, depth utilization should be evaluated explicitly, for example by replacing the depth input with depth from another scene, rather than inferring its contribution solely from task success. These recommendations apply to policies receiving depth as an image input and require validation on physical robotic systems.


\section*{ACKNOWLEDGMENT}
This work was supported by the National Science Foundation (NSF) through
Grant No.~2501703.

\balance
\bibliographystyle{IEEEtran}
\bibliography{references}

\end{document}